\documentclass[letterpaper, 10 pt, conference]{ieeeconf}  

\IEEEoverridecommandlockouts                              

\usepackage{graphicx}
\usepackage{subfigure} 
\usepackage{multirow}
\usepackage{siunitx}
\usepackage{booktabs} 
\usepackage[font=footnotesize]{caption}
\usepackage[utf8]{inputenc}
\usepackage[T1]{fontenc}
\usepackage{color}
\usepackage{caption}
\usepackage{siunitx}
\usepackage{amsmath}
\usepackage{threeparttable}

\usepackage{graphicx}
\usepackage{subcaption}
\usepackage{subfigure}

\usepackage{makecell}
\usepackage[linesnumbered,ruled,vlined]{algorithm2e}

\title{\LARGE \bf
TWC-SLAM: Multi-Agent Cooperative SLAM with Text Semantics and WiFi Features Integration for Similar Indoor Environments
}

\author{Chunyu Li$^{1}$, Shoubin Chen$^{1*}$, Dong Li$^{1}$, Weixing Xue$^{2}$ and Qingquan Li$^{1}$
\thanks{*This work was supported in part by the National Natural Science Foundation of China (42101445) and Director Foundation of Guangdong Laboratory of Artificial Intelligence and Digital Economy(SZ). }
\thanks{$^{*}$Corresponding author.}%
\thanks{$^{1}$Chunyu Li, Shoubin Chen, Dong Li and Qingquan Li are with the Guangdong Laboratory of Artificial Intelligence and Digital Economy (SZ), Shenzhen, China and Shenzhen University, Shenzhen 518060, China.
        {\tt\small lichunyu2020@email.szu.edu.cn, shoubin.chen@whu.edu.cn, doongli@ieee.org, liqq@szu.edu.cn}}%
\thanks{$^{2}$Weixing Xue is with the College of natural resources and environment, South China Agricultural University, Guangzhou 510642, China. {\tt\small wxxue@hotmail.com}}%
}

\begin{document}

\maketitle
\thispagestyle{empty}
\pagestyle{empty}

\begin{abstract}


Multi-agent cooperative SLAM often encounters challenges in similar indoor environments characterized by repetitive structures, such as corridors and rooms. These challenges can lead to significant inaccuracies in shared location identification when employing point cloud-based techniques. To mitigate these issues, we introduce TWC-SLAM, a multi-agent cooperative SLAM framework that integrates text semantics and WiFi signal features to enhance location identification and loop closure detection.
TWC-SLAM comprises a single-agent front-end odometry module based on FAST-LIO2, a location identification and loop closure detection module that leverages text semantics and WiFi features, and a global mapping module. The agents are equipped with sensors capable of capturing textual information and detecting WiFi signals. By correlating these data sources, TWC-SLAM establishes a common location, facilitating point cloud alignment across different agents' maps.
Furthermore, the system employs loop closure detection and optimization modules to achieve global optimization and cohesive mapping. We evaluated our approach using an indoor dataset featuring similar corridors, rooms, and text sign. The results demonstrate that TWC-SLAM significantly improves the performance of cooperative SLAM systems in complex environments with repetitive architectural features.

\end{abstract}

\section{INTRODUCTION}

Simultaneous Localization and Mapping (SLAM) technology is fundamental for enabling intelligent agents to autonomously navigate, localize themselves, and conduct 3D reconstructions in unknown environments. While effective SLAM solutions exist for straightforward scenarios, traditional single-agent SLAM approaches often encounter significant challenges in handling complex and similar scenes. In contrast, multi-agent cooperation can enhance performance in various applications, including inspection, search and rescue, infrastructure assessment, home services, warehouse logistics, and transportation, outperforming single-agent systems \cite{huang2021disco}.

\begin{figure}[!t]
    \centerline{\includegraphics[width=\columnwidth]{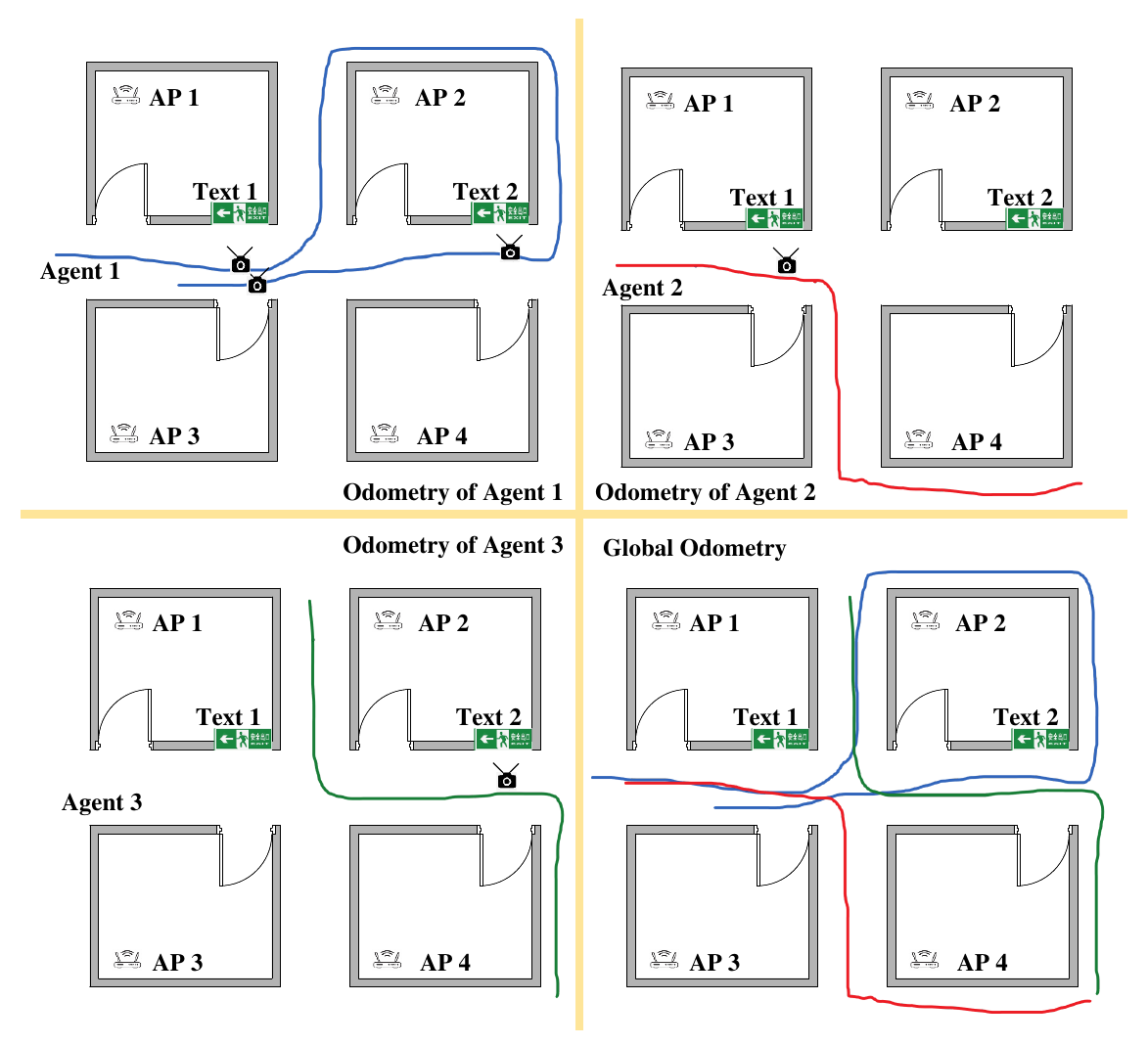}}
    \caption{\textbf{Schematic diagram of text semantic matching and WiFi feature matching}. Reliance on text semantic matching without complementary WiFi feature matching may induce localization inaccuracies. Specifically, two representative error patterns emerge: 1) In Agent 1's odometry trajectory, Text 2 could be erroneously associated with Text 1 as loop closure candidates; 2) Cross-agent confusion occurs when Text 2 captured by Agent 3's sensors is mismapped as Text 1. These erroneous associations propagate through the system, ultimately degrading global map consistency.}
    \label{fig_matching}
    \vspace{-10pt}
\end{figure}

LiDAR-based SLAM typically employs techniques such as Iterative Closest Point (ICP) \cite{besl1992method} and Normal Distributions Transform (NDT) \cite{chen2021ndt} to compute pose transformations between consecutive frame point clouds, subsequently aligning these point clouds within a unified coordinate system. By processing each frame sequentially, accurate odometry can be derived. Prominent methods in this domain include LOAM \cite{zhang2014loam}, NDT-LOAM \cite{chen2021ndt}, and more recent advancements like FAST-LIO \cite{xu2021fast} and LIO-SAM \cite{shan2020lio}. Multi-agent cooperative SLAM systems integrate a sub-map alignment module based on these foundational techniques, where alignment relies heavily on loop closure detection and location identification. Common approaches for location identification encompass Scan Context \cite{kim2018scan} and LiDAR-Iris \cite{wang2020lidar}, leveraging point cloud similarity. Furthermore, some methodologies incorporate image similarity, exemplified by LV-SLAM \cite{chen2021lidar}. Noteworthy classic multi-agent cooperative SLAM frameworks include DOOR-SLAM \cite{lajoie2020door} and DiSCo-SLAM \cite{huang2021disco}. However, these similarity-based methods often encounter matching inaccuracies in indoor environments characterized by abundant repetitive structures, such as narrow corridors and similar rooms.

This paper introduces \textbf{TWC-SLAM}, a cooperative LiDAR SLAM system that effectively integrates \textbf{text semantics and WiFi features, two of the most common characteristics in indoor scenarios.}
It is specifically designed for environments characterized by repetitive structures and textual elements, such as complex indoor spaces featuring long corridors and similar rooms. In such scenarios, traditional methods relying on point cloud and image similarity often encounter issues with false matches. Moreover, the prevalence of similar text signs in these indoor environments can lead to inaccuracies in conventional text-based SLAM approaches. The main contributions of this work are as follows:




\begin{enumerate}
    \item We propose a comprehensive multi-agent cooperative SLAM system. This approach pioneers the integration of text semantics with WiFi features to enable cooperative SLAM in multi-agent systems.

    \item We introduce a novel loop closure detection and a location recognition method that synergistically leverage text semantic matching and WiFi feature matching for enhanced cooperative SLAM performance.

    \item We constructed a multi-agent dataset and validated that the proposed approach demonstrates significant superiority over existing techniques in complex indoor environments with structurally homologous configurations and shared textual landmarks.
\end{enumerate}

\section{RELATED WORK}

\subsection{Cooperative SLAM}
Classic SLAM algorithms, such as LOAM \cite{zhang2014loam} and LeGO-LOAM \cite{shan2018lego}, extract corner and plane points as features to compute pose transformations for each point cloud frame. FAST-LIO \cite{xu2021fast} employs an extended Kalman filter to fuse point cloud data with inertial measurement unit (IMU) data, enabling high-precision mapping in challenging, fast-moving, and cluttered environments. FAST-LIO2 \cite{xu2022fast} further improves map maintenance techniques. LIO-SAM \cite{shan2020lio} formulates LiDAR-inertial odometry using a factor graph approach. R3LIVE \cite{lin2022r} comprises two subsystems: the LiDAR-inertial odometry (LIO) subsystem, which utilizes LiDAR and inertial sensor measurements to construct the geometric representation of the map, and the visual-inertial odometry (VIO) subsystem, which employs data from visual-inertial sensors to enhance the map's texture rendering.

Multi-agent cooperative SLAM extends the capabilities of single-agent SLAM systems. The SemanticCSLAM method \cite{li2024semanticcslam} integrates environmental landmarks with activity semantics and WiFi data to enhance multi-agent collaboration. CVI-SLAM \cite{karrer2018cvi} introduces a novel visual-inertial framework for centralized collaborative SLAM, offloading computationally intensive tasks by facilitating information sharing between agents and a central server. DOOR-SLAM \cite{lajoie2020door} employs point-to-point communication, negating the necessity for full connectivity among robots. This approach features two pivotal modules: a pose graph optimizer that integrates a distributed pairwise consistent measurement set maximization algorithm to mitigate false inter-robot loop closures, and a distributed SLAM front end capable of detecting inter-robot loop closures without the exchange of raw sensor data. Additionally, RDC-SLAM \cite{xie2021rdc} utilizes descriptor-based registration rather than traditional point cloud matching, applying eigenvalue-based segment descriptors to refine relative position estimates effectively.

\subsection{Location Identification}
The cornerstone of multi-agent cooperative SLAM lies in the identification of common locations by different agents and the effectiveness of loop closure detection. Traditional location identification methods in SLAM systems primarily rely on the similarity of point clouds. Global descriptor approaches, such as Scan Context \cite{kim2018scan}, Scan Context++ \cite{kim2021scan}, ISC \cite{wang2020intensity}, LocNet \cite{yin2018locnet}, and OverlapNet \cite{chen2020overlapnet}, establish correspondences by comparing the similarity of 2D images derived from 3D point clouds. Methods like SeqLPD \cite{liu2019seqlpd} and LPD-Net \cite{liu2019lpd} leverage PointNet \cite{qi2017pointnet} for feature extraction, utilizing NetVLAD \cite{arandjelovic2016netvlad} to identify similar features. SegMatch \cite{dube2017segmatch} presents a robust location recognition algorithm based on 3D fragment matching, while SegMap \cite{dube2018segmap} employs neural networks for feature extraction from point clouds. LV-SLAM \cite{chen2021lidar} further enhances loop closure detection by utilizing image similarity. In contrast, ASL-SLAM \cite{zhou2023asl} integrates activity semantics into the SLAM process. This method detects specific activities, such as traversing a speed bump or making a turn, to facilitate loop closure detection effectively.

\begin{figure*}[ht]
    \centering
    \resizebox{0.9\textwidth}{!}{\includegraphics{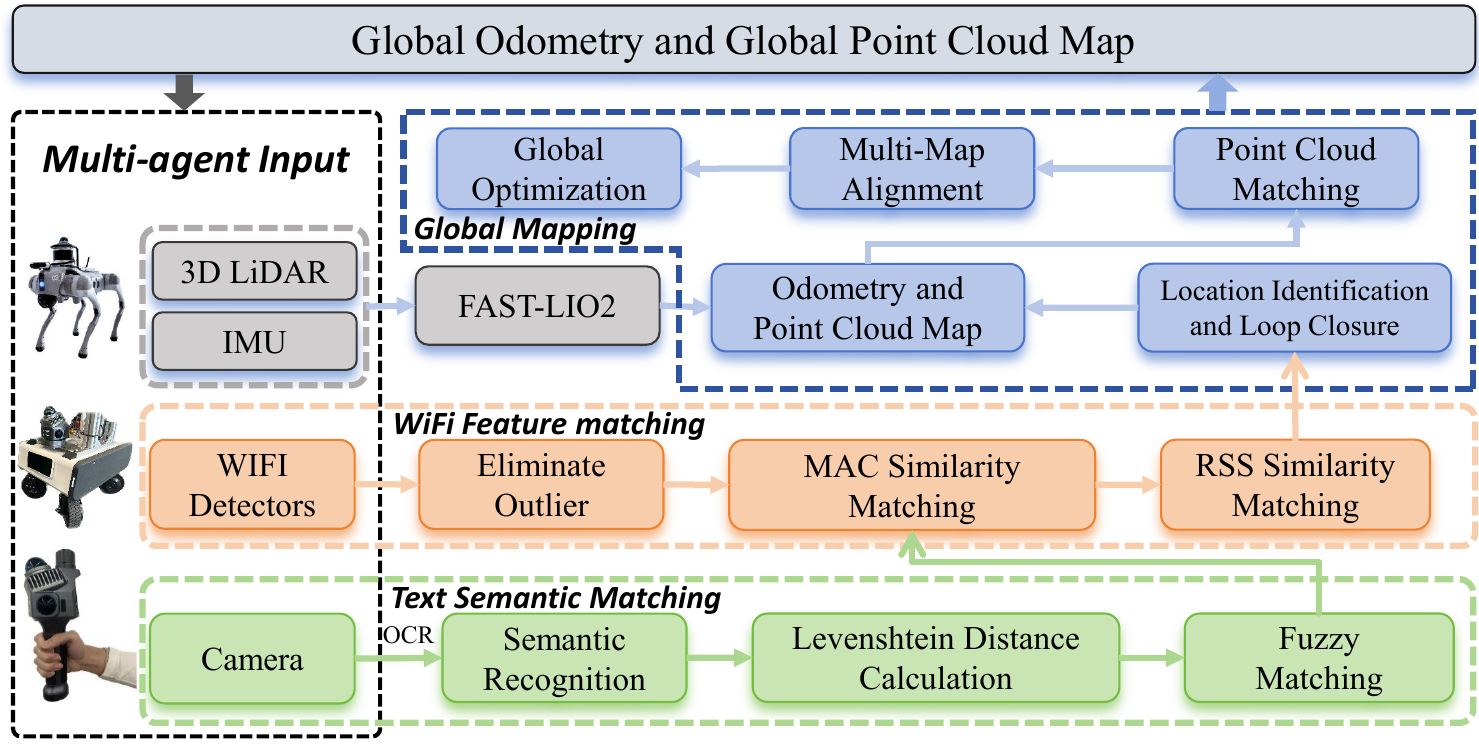}}
    \caption{\textbf{The overview of our proposed TWC-SLAM system}. The system comprises four main components: (1) Multi-Agent Input, (2) Text Semantic Matching, (3) WiFi Feature Matching, and (4) Global Mapping. Data is collected from diverse platforms, including handheld devices, wheeled robots, and legged robots, to enhance environmental adaptability.}
    \label{fig_system}
\end{figure*}

Utilizing textual information presents a viable approach for location identification within SLAM systems. TextSLAM \cite{li2020textslam,li2023textslam} enhances feature point detection by interpreting text boxes as planar surfaces. LP-SLAM \cite{zhang2023lp} leverages a large language model to extract textual information from the environment, subsequently employing this data for landmark identification and natural language navigation. TXSLAM \cite{tong2022txslam} conceptualizes text features as planes enriched with both texture and semantic information, leading to improved camera pose estimation through the tight coupling of these semantic planes.

In indoor positioning, Signal of Opportunity (SOP) is a prevalent technology, with WiFi signals serving as one of the primary modalities \cite{biswas2010wifi, shang2022overview, zafari2019survey}. SpotFi \cite{kotaru2015spotfi} employs a super-resolution algorithm, achieving a median positioning accuracy of 40 cm. WiDeep \cite{abbas2019wideep} integrates a stacked denoising autoencoder deep learning model with a probabilistic framework to mitigate noise in received WiFi signals and capture the intricate relationship between the WiFi signals detected by a device and its corresponding location. The method presented in \cite{salamah2016enhanced} leverages principal component analysis to enhance the performance of WiFi indoor positioning systems based on machine learning techniques while reducing computational costs. DLoc \cite{ayyalasomayajula2020deep} is a deep learning-based wireless positioning algorithm that enables WiFi devices to access an environmental map and estimate their positions relative to that map. Additionally, \cite{boonsriwai2013indoor} discusses the distinctions between multi-trilateration and WiFi fingerprinting concerning accuracy, computational complexity, and system resource consumption, proposing optimal configurations for these positioning algorithms to achieve more precise results.

\section{SYSTEM AND METHODS}

\subsection{System Overview}
As depicted in Fig. \ref{fig_system}, the proposed TWC-SLAM system comprises four core components: (1) The multi-agent frontend odometry module leverages the FAST-LIO2 algorithm \cite{xu2021fast} to compute odometry and generate point cloud maps for each agent. (2) The text semantic matching module extracts all textual semantics from the environment and applies fuzzy matching to identify pairs of similar text. (3) The WiFi feature matching module assesses the physical co-location of these matched text pairs by analyzing WiFi signal characteristics. (4) The global mapping module establishes shared reference points using correspondences derived from semantic text and WiFi features, subsequently performing loop closure detection and aligning trajectories and maps across agents to produce a globally consistent trajectory and map.

\subsection{Text Semantic Matching}
Before performing text semantic matching, we need to use Optical Character Recognition (OCR) technology to extract text semantics. In this paper, we select the PaddleOCR method \cite{zhou2017east, liao2020real}. Note that other OCR frameworks could also be used for this purpose. PaddleOCR demonstrates high accuracy in recognizing mixed combinations of text, letters, and numbers. It effectively identifies text semantics in both building and traffic environments. It writes the loss function $ L $ as follows:

\begin{equation}
    L = L_{s}+\lambda_{g}L_{g} \label{formula1}
\end{equation}
where $ L_{s} $ and $ L_{g} $ represent the score map and geometry losses respectively, and $ \lambda_{g} $ measures the importance between the two losses and set to 1.

\begin{equation}
    L_{s} = balanced-xent(Y_{1},Y_{2}) \label{formula2}
\end{equation}
where $ Y_{1} $ represents the predicted score map, and $ Y_{2} $ represents the ground truth.
Let all coordinate values of Q be an ordered set.

\begin{equation}
    C_{Q} = \{x_{1},y_{1},x_{2},y_{2}...x_{i},y_{i}\} \label{formula3}
\end{equation}

\begin{equation}
    L_{g} = min\sum_{c_{i}\in C_{Q}, c_{j} \in C_{Q*}} \frac{smoothed_{L1}(c_{i}-c_{i*})}{8 \times N_{Q}} \label{formula4}
\end{equation}
where the normalization term $ N_{Q} $ is the shorted edge length of the quadrangle.

After the text semantic recognition is completed, we need to detect the same text semantics. As shown in Fig. \ref{fig_matching}, we propose a multimodal matching algorithm as shown in Algorithm \ref{alg:multimodal}

The formula for the Levenshtein Distance \cite{li2023textslam} calculation method used in the calculation is as follows:

\begin{equation}
    S_{ij} =  \frac{max(\lvert S_{i} \rvert, \lvert S_{j} \rvert)-d(S_{i}, S_{j})}{max(\lvert S_{i} \rvert, \lvert S_{j} \rvert)} \label{formula5}
\end{equation}
where $ S_{ij} $ represents the text semantic similarity calculation score. $ \lvert S \rvert $ is the length of string, which represents the number of characters. $ d(S_{i}, S_{j}) $ is the distance between two strings, which represents the minimum operations to change one string $ S_{i} $ into another string $ S_{j} $, including deletion, insertion and substitution. We set thresholds $\alpha$. If $ S_{ij} \geq \alpha $, then we can consider the two texts to be the same.

After text semantic detection and matching, similar texts can be identified, which may suggest the same location. However, this cannot be confirmed with certainty, as identical texts may appear at different locations within the environment, such as emergency exit signs in various parts of a building. Therefore, we also need to employ a matching method based on WiFi features.

\subsection{WiFi Feature Matching}

As shown in Fig. \ref{fig_matching}, simply relying on the similarity of text semantics for location recognition may lead to errors. For example, if Robot A captures two identical texts, Text 1 and Text 2, the system might incorrectly match the location of Text 2 with the location of Text 1 during loop detection. Similarly, if Text 2 is captured by Robot 3, it could be mistakenly identified as Text 1, causing the location of Text 2 to be wrongly matched with the location of Text 1 when aligning the sub-point cloud map. Additionally, the location of Text 1 captured by Robot 2 may also be erroneously identified as the location of Text 2.

Therefore, it is necessary to perform text semantic matching and WiFi feature matching. The raw WiFi data primarily consists of three components: Access Point (AP) name, Media Access Control (MAC) address, and Received Signal Strength (RSS) value. The RSS value is related to the distance between the receiving device and the AP and generally follows the formula:

\begin{equation}
    RSS=P_{t}-K-10\zeta log_{10}d \label{formula6}
\end{equation}
where $ \zeta $ is called the path loss exponent, $ P_{t} $ is the transmit power, $ d $ is the distance from the AP to the receiving device , and $ K $ is a constant that depends on the environment and frequency. $ RSS $ can be used to calculate the distance between a mobile device and an AP. In general, the RSS value is also related to the environment. For example, wall occlusion will weaken the RSS value. In current buildings, signals from multiple APs can generally be received, so the RSS values of multip APs can be used to calculate the feature together. Because an AP name can correspond to multiple different MAC addresses, we use the MAC address as the feature instead of the AP name.

At the same location, the RSS value is not a fixed value. The RSS standard deviation can be calculated using the following formula:

\begin{equation}
    R=\sqrt{(X_{1}-\overline{X})^2+(X_{2}-\overline{X})^2+\dots(X_{n}-\overline{X})^2} \label{formula7}
\end{equation}

\begin{equation}
    \overline{X}=\frac{X_{1}+X_{2}+\dots X_{n}}{n} \label{formula8}
\end{equation}
where $ X_{1}, X_{2}, \dots ,X_{n} $ represents the RSS values from the same MAC detected within a period of time.

Then, we removed the RSS values that were larger than the standard deviation $ R $, and averaged the remaining RSS values to get $ RSS^* $. Then we can get the different MAC addresses and the corresponding RSS values detected in a certain location, expressed as $ (MAC_{1}:RSS_{1}^*, MAC_{2}:RSS_{2}^*, \dots, MAC_{n}:RSS_{n}^*) $.

To compare the WiFi feature of the two locations, we can compare the MAC and $ RSS^* $ of the two locations. We first need to compare the MAC addresses of the two locations. This can be calculated using the following formula:

\begin{equation}
    M_{s}=\frac{M_{c}}{max(M_{i}, M_{j})} \label{formula9}
\end{equation}
where $ M_{i} $ and $ M_{j} $ represent the number of the MAC address appears in locations $ i $ and $ j $, and $ M_{c} $ represents the number of times the same MAC address appears together. The value of $ M_{s} $ needs to be greater than the threshold $ \beta $ before the next step is executed.

Taking out the same MAC of the two frames and corresponding $ RSS^* $, we can calculate the overall WiFi feature similarity of the two locations by the following formula (\ref{formula10}).

\begin{equation}
    d(L_{i}, L_{j}) = \sqrt{
    \sum_{k=1}^{n} (RSS_{ik}^* - RSS_{jk}^*)^2 \label{formula10}
    }
\end{equation}
where $ d(L_{i},L_{j}) $ represents the similarity of the $ RSS^* $ of two locations $ i $ and $ j $. $ RSS_{in}^* $ and $ RSS_{jn}^* $ represent the $ RSS^* $ value of the MAC corresponding to locations $ i $ and $ j $. The value of $ d(L_{i},L_{j}) $ needs to be greater than the threshold $ \gamma $.

In summary, the WiFi feature matching task requires both indicators to be met. The first indicator is the similarity indicator $ M_{s} $ of the detected MAC address, which needs to be greater than the threshold $ \beta $. The second indicator is the similarity indicator $ d(L_{i},L_{j}) $ of the RSS value, which needs to be greater than the threshold $ \gamma $. 

\begin{algorithm}[ht]
\caption{Multi-modal Location Recognition}
\label{alg:multimodal}
\KwIn{Location $i$ data: Text$_i$, $\{(MAC_1: \{RSS_1\}), ..., (MAC_m: \{RSS_m\})\}$ \\
Location $j$ data: Text$_j$, $\{(MAC_1: \{RSS_1\}), ..., (MAC_n: \{RSS_n\})\}$ \\
Thresholds $\alpha$, $\beta$, $\gamma$}
\KwOut{Matching decision (True/False)}

\BlankLine
\textbf{Step 1: Text Semantic Matching}
Compute Levenshtein similarity score:
$S_{ij} = \frac{\max(|S_i|, |S_j|) - d(S_i,S_j)}{\max(|S_i|, |S_j|)}$ \tcp*{Eq.\ref{formula5}}

\If{$S_{ij} < \alpha$}{
    \Return False \tcp*{Text similarity below threshold}
}

\BlankLine
\textbf{Step 2: WiFi Feature Matching}
\ForEach{location $k \in \{i,j\}$}{
    \ForEach{MAC address in location $k$}{
        $\overline{X} \leftarrow \frac{1}{n}\sum_{i=1}^n X_i$ \tcp*{RSS mean (Eq.\ref{formula8})}
        $R \leftarrow \sqrt{\sum_{i=1}^n (X_i-\overline{X})^2}$ \tcp*{Std dev (Eq.\ref{formula7})}
        Filter RSS values where $|X-\overline{X}| \leq R$\;
        $RSS^*_k \leftarrow$ average of filtered values \tcp*{Using Eq.\ref{formula6}}
    }
}

Compute MAC similarity:
$M_s = \frac{M_c}{\max(M_i,M_j)}$ \tcp*{Eq.\ref{formula9}}

\If{$M_s < \beta$}{
    \Return False \tcp*{MAC similarity too low}
}

Compute RSS distance:
$d(L_i,L_j) = \sqrt{\sum_{k=1}^p (RSS^*_{ik}-RSS^*_{jk})^2}$ \tcp*{Eq.\ref{formula10}}

\If{$d(L_i,L_j) \geq \gamma$}{
    \Return True \tcp*{Both modalities match}
}
\Else{
    \Return False \tcp*{RSS similarity too low}
}
\end{algorithm}

\subsection{Global Mapping}
After matching the text semantics and WiFi features, it is possible to identify the same position. Single-agent SLAM can then perform loop closure detection based on this position identification. As shown in Fig.\ref{fig_matching}, Robot 1 first detects texts 1 and 2, and then uses WiFi features to distinguish between them for loop closure detection.

We can also identify the same locations visited by different agents, and then execute the global mapping module. This module comprises three components: point cloud matching, multi-map alignment, and global optimization.

Once a common location is correctly identified, the coordinate transformation between the sub-point cloud maps needs to be calculated using a point cloud matching method, such as the Iterative Closest Point (ICP) method.

Define two point cloud sets $ P = \{p_{1}, p_{2}, ..., p_{n}\} $ and $ Q = \{q_{1}, q_{2}, ..., q_{n}\} $. Where $ P $ is the reference point cloud set and $ Q $ is the data point cloud set. Point cloud registration requires a series of rotations and translations from $ Q $ to the target point cloud $ P $. In fact, the two point cloud sets are not exactly the same, so it is often impossible to get accurate $ R $ and $ T $ to make the two completely coincide, but the transformed point cloud $ Q $ can be made as close to $ P $ as possible, so the problem can be described as:

\begin{equation}
    P_{t}^{i} = R \cdot P_{s}^{i}+t \label{formula11}
\end{equation}

\begin{equation}
    F(R,T) = \frac{1}{n} \sum_{i=1}^{n} \left| P_{t}^{i}-R \cdot P_{s}^{i}-t \right| ^{2} \label{formula12}
\end{equation}

Since the corresponding points are initially unknown, we assume initial values of the rotation and translation matrices and transform the source point cloud by these initial matrices to obtain the transformed point cloud. We then compare this transformed point cloud with the target point cloud. If the distance between corresponding points in the two point clouds is below a certain threshold, we consider these points to be corresponding. This process is known as finding the nearest point pairs. Using these corresponding points, we can estimate the rotation matrix 
$ R $ and the translation vector $ T $. This estimation is framed as a least squares problem, which is converted into an optimization problem where $ R $ and $ T $ are determined by minimizing the sum of the squared errors of the objective function.

To calculate the coordinate transformation between two sub-point cloud maps, only two frames of point clouds are required. If multiple common positions are identified between the two sub-maps, then point clouds from multiple frames can be used for calculation, leading to more accurate results. Once all sub-point cloud maps are transformed into a unified coordinate system, the alignment of the multi-map task is completed.

\section{EXPERIMENTS AND RESULTS}
\subsection{Dataset and Implementation Details}

\begin{figure}[!t]
    \centering{
        \subfigure[]{\includegraphics[width=0.4\columnwidth]{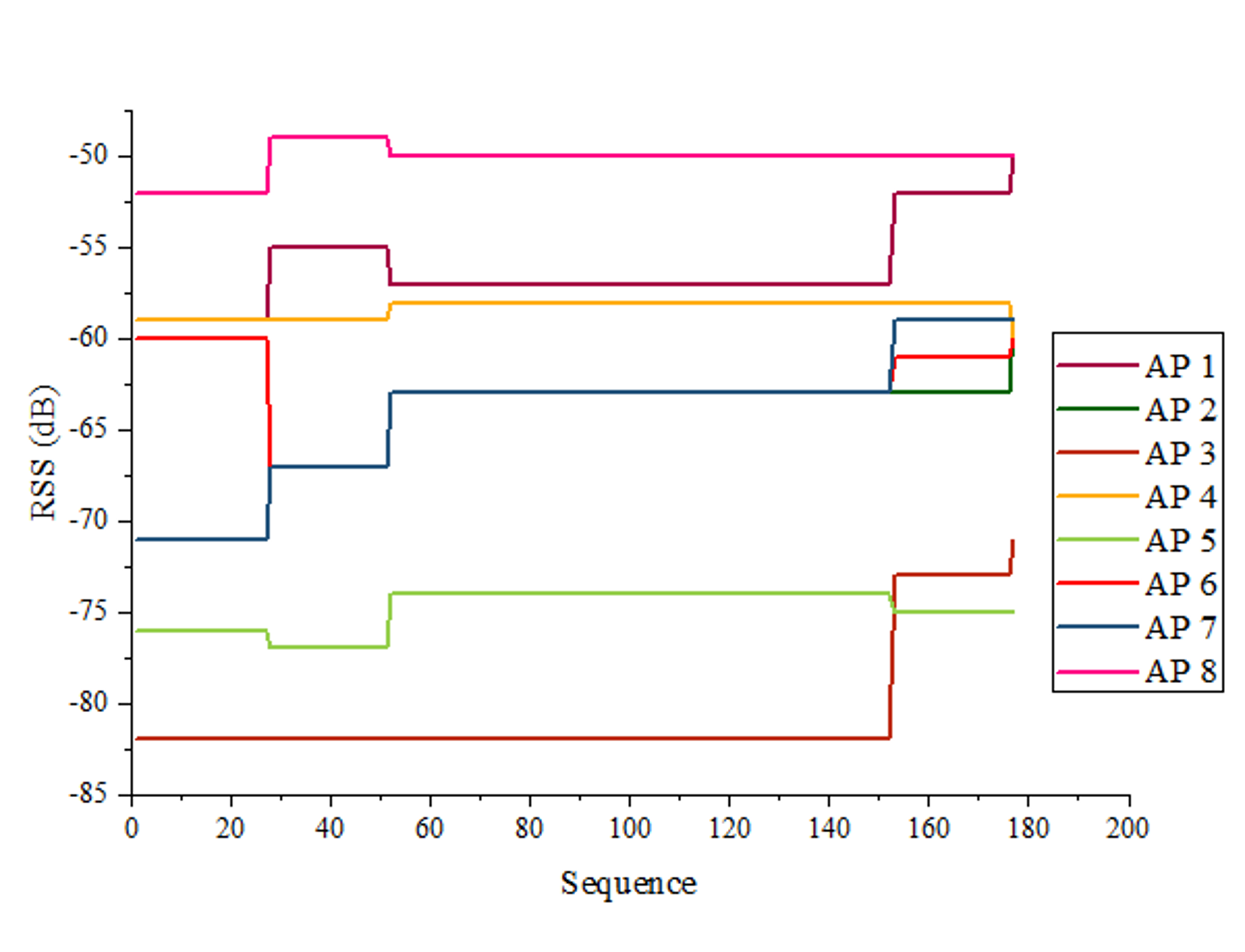}}
        \subfigure[]{\includegraphics[width=0.2\columnwidth]{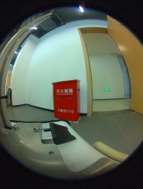}}
        \subfigure[]{\includegraphics[width=0.2\columnwidth]{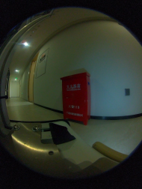}}}
    \caption{\textbf{WiFi and text conditions in the experimental scenario.} Figure (a) is an example of some WiFi information, while Figures (b) and (c) demonstrate identical text semantics positioned at distinct locations.}
    \label{fig:conditions}
\end{figure}

In our experiment, we utilized the MetaCam EDU handheld device, as well as wheeled and legged robots, for comprehensive data collection, employing a Mid360 LiDAR as the primary sensor. Furthermore, WiFi fingerprint data was gathered using a mobile phone.  


Our experimental validation was conducted across two distinct floors of the Guangming Laboratory. These selected environments present various architectural challenges, including repetitive room configurations, elongated corridor structures, spatially distributed duplicate textual markings, and strategically positioned WiFi access points. Together, these factors represent complex indoor localization scenarios.

As shown in Fig. \ref{fig:conditions}, the agent successfully detects signals from more than eight distinct APs with varying RSS values in these scenarios. Notably, the experimental environment contains multiple identical text elements positioned at distinct locations, which effectively validates the technical merits of our proposed approach.

\subsection{Same Location Recognition}
The core challenge in cooperative SLAM systems lies in effectively identifying overlapping locations traversed by multiple agents, which serves as the prerequisite for integrating their respective trajectories and environmental maps into a unified coordinate framework. Our proposed methodology enhances location recognition accuracy through a multimodal approach that combines text semantic analysis with WiFi fingerprint characterization, enabling robust cross-agent map alignment.

As shown in Table \ref{table_location}, we conducted controlled experiments by isolating text semantics and WiFi fingerprints as independent variables to analyze their individual and combined effects on location recognition performance. The evaluation metric for common location detection of different trajectories is formally defined as:

\textbf{Only-text-based Location Precision}:
\begin{equation}
        P_{\text{text}} = \frac{|\{ \texttt{Correct\ text\ pairs} \}|}{|\{ \texttt{Candidate text pairs} \}|} \times 100\%
\end{equation}
where the numerator represents verified co-located text pairs across trajectories, and the denominator denotes all semantically matched text pairs.

\textbf{Only-text-based Location Recall}:
\begin{equation}
        R_{text} = \frac{|\{ \texttt{Detected true positives} \}|}{|\{ \texttt{Ground truth co-located pairs} \}|} \times 100\%
\end{equation}
where the numerator represents the number of pairs of identical text pairs detected across tracks that are actually located at the same location, and the denominator represents the number of identical text pairs across tracks in reality.
    
\textbf{Only-WiFi-based Location Precision}:
\begin{equation}
        P_{\text{WiFi}} = \frac{|\{ \texttt{Correct WiFi pairs} \}|}{|\{ \texttt{Candidate WiFi pairs} \}|} \times 100\%
\end{equation}
where the numerator represents the number of correct identical locations detected by wifi similarity, and the denominator represents the number of all identical locations detected by wifi similarity.
    
\textbf{Only-WiFi-based Location Recall}:
\begin{equation}
        R_{WiFi} = \frac{|\{ \texttt{Detected true location} \}|}{|\{ \texttt{All same locations} \}|} \times 100\%
\end{equation}
where the numerator represents the correct number of identical locations detected using the WiFi fingerprint method, and the denominator represents the number of identical locations in reality.

\begin{table*}
    \centering
    \caption{Results of The Same Location Recognition}
    \label{table_location}
    \setlength{\tabcolsep}{3pt} 
    \begin{tabular}{@{}c *{9}{cc}@{}} 
        \hline
        \multirow{2}{*}{\makecell{Scene\\ (\#)}} 
        & \multicolumn{6}{c}{Text-only (Threshold $\alpha$)} 
        & \multicolumn{6}{c}{WiFi-only (Threshold $\beta, \gamma$)}
        & \multicolumn{6}{c}{Text-WiFi-based (Threshold $\alpha, \beta, \gamma$)} \\
        \cmidrule(lr){2-7} 
        \cmidrule(lr){8-13} 
        \cmidrule(l){14-19} 
        & \multicolumn{2}{c}{$\alpha=0.5$} 
        & \multicolumn{2}{c}{$\alpha=0.8$} 
        & \multicolumn{2}{c}{$\alpha=1.0$} 
        & \multicolumn{2}{c}{$\beta, \gamma=0.5$} 
        & \multicolumn{2}{c}{$\beta, \gamma=0.8$} 
        & \multicolumn{2}{c}{$\beta, \gamma=1.0$} 
        & \multicolumn{2}{c}{$\alpha=0.5$ $\beta, \gamma=0.8$} 
        & \multicolumn{2}{c}{$\alpha=0.8$ $\beta, \gamma=0.8$} 
        & \multicolumn{2}{c}{$\alpha=0.5$ $\beta, \gamma=1.0$} \\
        \cmidrule(lr){2-3} 
        \cmidrule(lr){4-5} 
        \cmidrule(lr){6-7} 
        \cmidrule(lr){8-9} 
        \cmidrule(lr){10-11} 
        \cmidrule(lr){12-13}
        \cmidrule(lr){14-15} 
        \cmidrule(lr){16-17} 
        \cmidrule(l){18-19} 
        & P & R & P & R & P & R & P & R & P & R & P & R & P & R & P & R & P & R \\
        \hline
        \texttt{\#01} & 35\% &89\% &52\% &78\% &78\% &31\% &45\% &86\% &62\% &72\% &68\% &70\% &65\% &92\% &95\% &90\% &96\% &65\% \\
        \texttt{\#02} & 41\% &75\% &61\% &76\% &88\% &45\% &51\% &81\% &70\% &76\% &87\% &40\% &54\% &89\% &82\% &93\% &84\% &68\% \\
        \hline
    \end{tabular}
\end{table*}

The analysis of Table \ref{table_location} reveals that text-semantics-based location recognition achieves reasonable accuracy only when employing an extreme threshold. However, practical implementations face inherent challenges: variations in lighting conditions, camera angles, and agent mobility during text image capture may yield inconsistent semantic detection outcomes even for identical textual content, consequently resulting in unsatisfactory recall rates. Meanwhile, spatial variations in WiFi signal strength typically exhibit limited discriminative power within confined areas, and WiFi-exclusive position recognition tends to produce incorrect matches with adjacent locations, thus compromising positioning accuracy. This dual limitation necessitates the integration of textual semantics with WiFi fingerprint. Through empirical analysis, we established optimal operational thresholds at $\alpha=0.8$ and $\beta, \gamma=0.8$. This parameter combination demonstrates balanced performance by simultaneously optimizing both precision and recall metrics in same location recognition tasks.

\subsection{Global Mapping}
\begin{figure}[!t]
    \centering{
        \includegraphics[width=1.0\columnwidth]{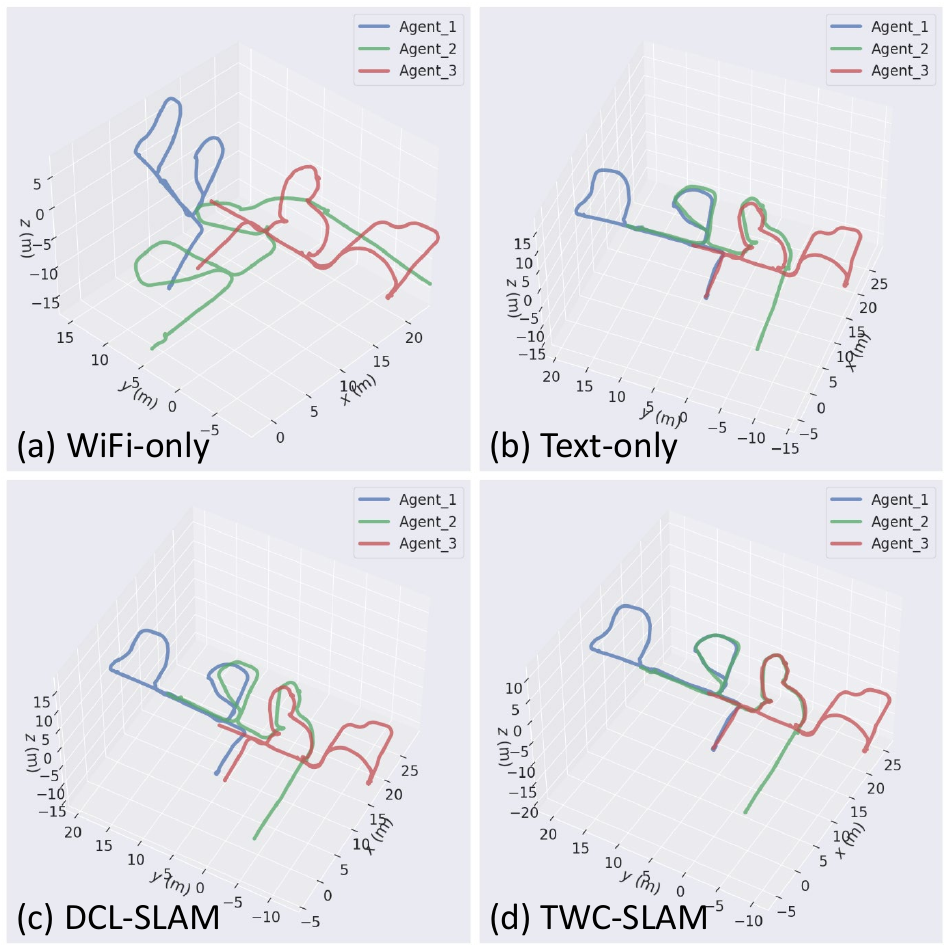}}
        
    \caption{\textbf{Comparison of trajectory results for Scene \texttt{\#01}}. The blue odometry in (a) has completely deviated, and the odometry in (b) and (c) have also deviated.}
    \label{fig_4f}
\end{figure}

\begin{figure}[!t]
    \centering{
        \includegraphics[width=1.0\columnwidth]{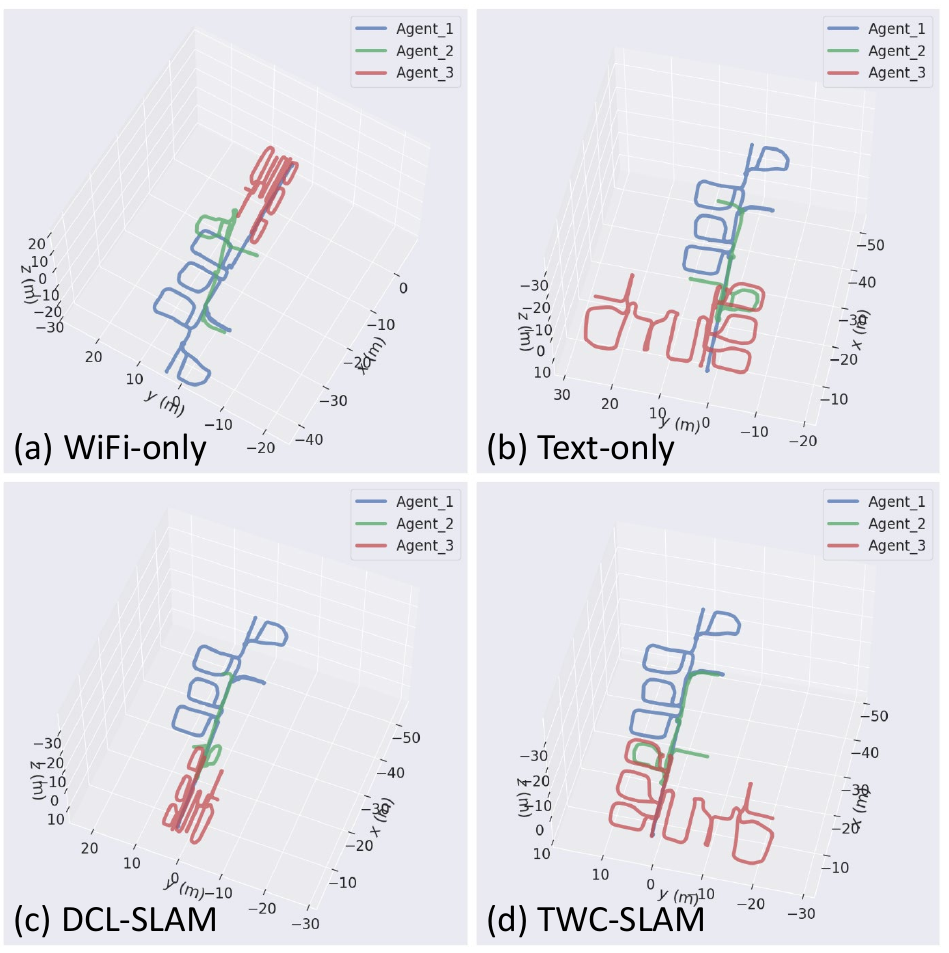}}
    \caption{\textbf{Comparison of trajectory results for Scene \texttt{\#02}}. The red odometry in (a) and (c) have completely deviated, and the odometry in (b) has deviated.}
    \label{fig_5f}
\end{figure}

\begin{figure}[!t]
    \centering{
        \includegraphics[width=0.6\columnwidth]{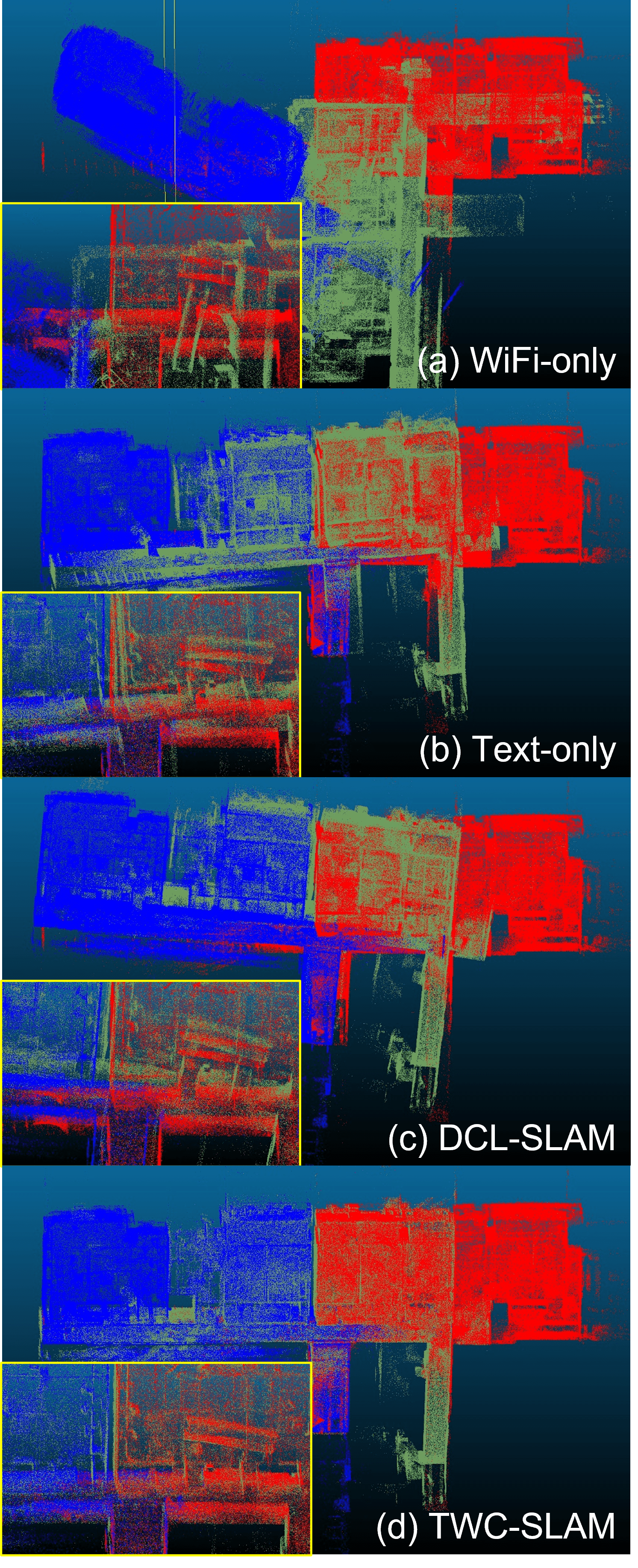}}
    \caption{\textbf{Comparison of point cloud results for Scene \texttt{\#01}}. The chromatic attributes of the point cloud maintain precise alignment with the trajectory's coloration scheme presented in Fig. \ref{fig_4f}.}
    \label{fig_point}
\end{figure}

\begin{table*}[!ht]
\centering
\begin{threeparttable} 
\caption{Comparison of EPE Results (units in meters.)}
\label{table_results}
\begin{tabular}{lccccc}
\toprule
Scene (\#) & Travel Distance\tnote{a} & DCL-SLAM & TWC-SLAM (text) & TWC-SLAM (WiFi) & \textbf{TWC-SLAM} \\
\midrule
\texttt{\#01} & 265.32 & 1.69 & 0.35 & 3.16 & \textbf{0.21} \\
\texttt{\#02} & 463.91 & 1.41 & 1.67 & 1.56 & \textbf{0.16}\\
\bottomrule
\end{tabular}
\begin{tablenotes}
\item[a] Travel distance refers to the sum of the lengths of all trajectories.
\end{tablenotes}
\end{threeparttable}
\end{table*}

As shown in Table \ref{table_results}, we compared our method with three approaches: DCL-SLAM~\cite{zhong2023dcl}, a cooperative SLAM method employing LiDAR-Iris~\cite{wang2020lidar} for point cloud similarity, and our method variants using only text semantics or WiFi for location recognition. For the text-only semantic approach, we selected the threshold $\alpha=1.0$ through empirical validation as it achieved optimal recognition accuracy. In the WiFi-based method, the parameter configuration $\beta, \gamma=1.0$ was adopted to maximize positioning precision based on systematic evaluations. Given the absence of GPS signals and ground truth data in indoor environments, we employed the start-end closure method to calculate and compare End Point Error (EPE). 

As illustrated in Fig. \ref{fig_4f} and Fig. \ref{fig_point}, Scene \texttt{\#01} presents four geometrically analogous rooms and long corridor that induce structural ambiguity. This configuration misled DCL-SLAM's place recognition module, accumulating 1.69m positioning drift through erroneous loop closures. The repetitive fire extinguisher cabinets further introduced 0.35m deviation in text-only semantic localization due to textual aliasing. While WiFi-based localization completely failed from multipath interference in this environment, our method successfully distinguished distinct rooms through combining text and WiFi, achieving superior positioning accuracy (0.21m). TWC-SLAM demonstrates minimal ghosting artifacts in point cloud reconstruction, with distinct pipeline contours observable along the ceiling region. Comparative analyses reveal noticeable ghosting effects in both DCL-SLAM and Text-only outputs, while the WiFi-only mapping framework fails to produce viable structural representations.

In the more challenging Scene \texttt{\#02} shown in Fig. \ref{fig_5f}, three complex trajectories involve repeated entries/exits to structurally similar rooms. The geometric ambiguity caused DCL-SLAM's point cloud matching to accumulate 1.41m error, while duplicated emergency exit signs misled text-only localization to 1.67m deviation. WiFi positioning remained unreliable (1.56m error) due to inherent signal noise. Our multimodal approach maintained robust performance (0.16m error) by effectively resolving both structural and text ambiguities through cross-modal feature fusion. Comparative analysis of methodologies in Scene \texttt{\#02} is discernible through trajectory visualization, therefore rendering point cloud representation unnecessary.

\section{CONCLUSIONS}
We propose a multi-agent cooperative SLAM system specifically tailored for challenging environments featuring structural similarities and textual ambiguity. Our methodology integrates text semantics with WiFi fingerprint correlations to establish inter-agent correspondences, subsequently performing joint optimization of sub-odometry trajectories and sub-maps for global consistency. To validate the system, we have constructed a novel multi-agent dataset capturing representative indoor scenarios with repetitive architectural patterns and text-rich surfaces. Experimental results reveal significant performance advantages over existing solutions: achieving 88\% higher precision than point cloud-based methods, 82\% improvement over text-only methods, and 92\% enhancement compared with WiFi-only methods in cooperative SLAM tasks.

\bibliographystyle{IEEEtran}
\bibliography{main}
\end{document}